\documentclass[runningheads]{llncs}

\usepackage[T1]{fontenc}
\usepackage{hyperref}
\usepackage{graphicx}
\usepackage{mdframed}
\usepackage{makecell}
\usepackage{xcolor}
\definecolor{pink}{rgb}{0.95, 0.8, 0.96}
\definecolor{pinkd}{rgb}{0.95, 0.4, 0.96}
\usepackage[textsize=tiny,backgroundcolor=pink,linecolor=pinkd]{todonotes}

\newcommand{\entity}[1]{\textsf{\small #1}}
\newcommand{\rel}[1]{\textsl{\small #1}}

\newcommand{\triple}[3]{(\entity{#1}, \rel{#2}, \entity{#3})}

\newcommand{\hitone}{\texttt{\small Hits@1$^\uparrow$}}

\newcommand{\hitten}{\texttt{\small Hits@10$^\uparrow$}}

\newcommand{\mr}{\texttt{\small MR$^\downarrow$}}
\newcommand{\mrr}{\texttt{\small MRR$^\uparrow$}}

\newcommand{\map}{\texttt{\small MAP@100$^\uparrow$}}

\newcommand{\phund}{\texttt{\small P@100$^\uparrow$}}

\newcommand{\eat}[1]{}
\newcommand{\gotoTR}[1]{}

\newcommand{\origunderscore}{}
\let\origunderscore\_
\renewcommand{\_}{\allowbreak\origunderscore}

\begin{document}

\title{On Large-Scale Evaluation of Embedding Models for Knowledge Graph Completion}

\titlerunning{On Large-scale Evaluation of Embedding Models for KG Completion}

\author{Nasim Shirvani-Mahdavi\orcidID{0009-0006-2733-2242} \and
Farahnaz Akrami \and \\
Chengkai Li\orcidID{0000-0002-1724-8278}}

\authorrunning{N. Shirvani-Mahdavi et al.}

\institute{University of Texas at Arlington, Arlington TX, 76019, USA \\
\email{\{nasim.shirvanimahdavi2, farahnaz.akrami\}@mavs.uta.edu}, \email{cli@uta.edu}}

\maketitle   
\begin{abstract} Knowledge graph embedding (KGE) models are extensively studied for knowledge graph completion, yet their evaluation remains constrained by unrealistic benchmarks. Standard evaluation metrics rely on the closed-world assumption, which penalizes models for correctly predicting missing triples, contradicting the fundamental goals of link prediction. These metrics often compress accuracy assessment into a single value, obscuring models' specific strengths and weaknesses. The prevailing evaluation protocol, link prediction, operates under the unrealistic assumption that an entity's properties, for which values are to be predicted, are known in advance. While alternative protocols such as property prediction, entity-pair ranking, and triple classification address some of these limitations, they remain underutilized. 
Moreover, commonly used datasets are either faulty or too small to reflect real-world data. Few studies examine the role of mediator nodes, which are essential for modeling n-ary relationships, or investigate model performance variation across domains. This paper conducts a comprehensive evaluation of four representative KGE models on large-scale datasets FB-CVT-REV and FB+CVT-REV. Our analysis reveals critical insights, including substantial performance variations between small and large datasets, both in relative rankings and absolute metrics, systematic overestimation of model capabilities when n-ary relations are binarized, and fundamental limitations in current evaluation protocols and metrics.

\end{abstract}
\section{Introduction}\label{sec:intro} 
Knowledge graphs (KGs), such as Freebase~\cite{freebase} and Wikidata~\cite{wikidata}, represent real-world information in the form of triples, denoted as \triple{head entity (subject)}{relation}{tail entity (object)}. KGs are indispensable resources for a wide range of artificial intelligence applications~\cite{ji2021survey}, such as question answering~\cite{questionanswering} and recommender systems~\cite{recommendersurvey}. 
Despite their size and utility, KGs are inherently incomplete, which can hinder their effectiveness in practical applications. Researchers have developed numerous automatic knowledge graph completion techniques to address this limitation. Among these approaches, knowledge graph \emph{embedding} (KGE) models~\cite{cai2018comprehensive} have been extensively studied. These models aim to infer missing information by leveraging the latent representations of entities and relations. 

This paper presents a comprehensive large-scale evaluation of KGE models for KG completion. Although these models have been extensively evaluated to assess their efficacy and applicability in real-world applications, we demonstrate fundamental limitations in current evaluation methodologies that result in inaccurate assessments of model performance. Our investigation focuses particularly on large-scale analysis, examining two main interrelated issues: 1) shortcomings of standard rank-based evaluation metrics and 2) inherent weaknesses in evaluation protocols. Our findings reveal that these methodological limitations substantially affect how KGE models are understood and compared. We also examine how data representation choices affect performance outcomes by using two datasets that employ different data modeling approaches in all our experiments.

Many existing evaluations of KGE models rely on datasets, such as FB15k-237~\cite{bordes2013translating} and WN18RR~\cite{ConvE}, both of which exhibit several limitations. First, they fail to capture n-ary relationships among entities, which are modeled using mediator nodes (also called CVT nodes) in the Freebase data dump~\cite{pellissier2016freebase} and statement nodes in Wikidata. Our earlier findings~\cite{shirvani2023comprehensive} indicate that the conversion of n-ary relationships to binary ones, as implemented in FB15k-237, leads to substantial overestimation of KGE models' performance. Second, these datasets are considerably smaller than real-world datasets. Experiment results from previous studies~\cite{shirvani2023comprehensive,dgl-ke} reveal significant differences in the performance of models between small-scale and large-scale datasets, with relative performance rankings among models also varying based on dataset size. Our experiments are conducted on two large-scale datasets, FB-CVT-REV and FB+CVT-REV~\cite{shirvani2023comprehensive}. While FB+CVT-REV represents n-ary relationships using CVT nodes, FB-CVT-REV converts n-ary relationships into binary ones, similar to FB15k-237. This distinction allows us to assess how KGE model performance varies when data is modeled differently. These datasets, derived from Freebase, are large-scale, represent real-world scenarios, and do not include the data redundancy that exists in FB15k. A discussion of these two datasets and other benchmarks is provided in Section~\ref{sec:background}.

(1) \textit{Shortcomings of standard rank-based evaluation metrics}: KG completion methods aim to identify missing but correct triples in a KG. However, current evaluation metrics paradoxically penalize models for correctly predicting triples that are missing from the evaluation dataset. Standard ranking-based metrics such as Mean Rank (MR) and Mean Reciprocal Rank (MRR)~\cite{yang2022rethinking} operate under the \emph{closed-world assumption}, treating all predictions outside the KG as incorrect. This contradicts the \emph{open-world assumption}, which acknowledges that absent facts are not necessarily false~\cite{reiter1981closed}. Furthermore, each of these metrics aggregates performance across all relations and triples into a single value, obscuring models' strengths and weaknesses on different relations. In Section~\ref{sec:issues}, we explore the shortcomings of existing evaluation metrics with supporting experimental evidence discussed in Section~\ref{sec:experiments}. We illustrate how these shortcomings impact models' performance and propose refinements to improve evaluation reliability. 

(2) \textit{Inherent weaknesses in evaluation protocols}: 
The standard evaluation protocol for KGE models, \emph{link prediction}~\cite{bordes2013translating}, evaluates models by predicting the missing \entity{h} in triple \triple{?}{r}{t} or missing \entity{t} in \triple{h}{r}{?}. It assesses whether a model prioritizes correct answers over incorrect ones. However, as Wang et al.~\cite{wang-etal-2019-evaluating} pointed out, this approach can be misleading because it does not verify whether correct triples are ranked better than all type of false and nonsensical ones. For example, for a test triple \triple{James Ivory}{director/film}{A Room With A View}, evaluation by link prediction does not rank the nonsensical triple \triple{A Room With A View}{director/film}{James Ivory}. Wang et al.~\cite{wang-etal-2019-evaluating} showed that models often assign high scores to such invalid triples. To assess how models rank all possible triples, even the nonsensical ones, for a given relation, we performed the \emph{entity-pair ranking} evaluation protocol on the two datasets mentioned above. Another limitation of link prediction is its assumption that an entity's properties are already known, focusing only on retrieving the correct object or subject entities. In practice, determining whether a property applies to an entity is itself a challenge. To address this gap, we introduce a new evaluation protocol, \emph{property prediction}. A third alternative evaluation protocol to link prediction, \emph{triple classification}~\cite{TransH,socher2013reasoning}, assesses whether a given triple is a true or false fact. This requires negative triples in the test and validation sets. Traditionally, these negative triples are generated by randomly replacing a head or tail entity, an approach that does not sufficiently challenge the models~\cite{socher2013reasoning,TransH}. Inspired by \emph{hard negatives} suggested by Pezeshkpour et al.~\cite{pezeshkpour2020revisiting} and Safavi et al.~\cite{codex}, we propose focusing on generating \emph{type-consistent} negatives using the Freebase type system as models generally excel at identifying entity types, making \emph{type-inconsistent} negatives easier to classify and providing less meaningful evaluation challenges. Further details on the limitations of link prediction can be found in Section~\ref{sec:issues}, while experiments involving alternative evaluation protocols are presented in Section~\ref{sec:experiments}.

\vspace{0.07 cm} 
In summary, this paper makes the following contributions:\vspace{-0.1 cm} 
\begin{itemize}
    \item We present a comprehensive large-scale evaluation of KGE models on FB-CVT-REV and FB+CVT-REV datasets.
    \item We provide quantitative results indicating that metrics based on the closed-world assumption underestimate models' accuracy.
    \item We develop comprehensive global metrics revealing hidden performance patterns, including macro-averaging by relation or domain, analysis of left and right entity link predictions, and quantitative evidence showing how n-ary to binary conversion leads to overestimation of model performance.
    \item We highlight the limitations of link prediction as an evaluation protocol for KGE models, 
    and demonstrate how dataset size affects model performance and relative rankings among different KGE approaches.
    \item We introduce a new evaluation protocol, property prediction, and evaluate the KGE models on this protocol, entity-pair ranking, and triple classification as alternatives to link prediction.
\end{itemize}

\vspace{-0.1 cm}
\noindent All the scripts and datasets used in our experiments, and additional details on the experimental setup, are publicly available at \sloppy 
\href{https://github.com/idirlab/largeKGEeval}{https://github.com/idirlab/largeKGEeval}.

\vspace{-0.4 cm}
\section{Related work}\label{sec:related}
\vspace{-0.3 cm}
Prior research has investigated various aspects of KGE evaluation. Several studies have explored the impact of training strategies on  KGE models' performance~\cite{ali2021bringing}.  Evaluation metrics for KGE models have been extensively studied. For instance, Hoyt et al.~\cite{hoyt2022unified} proposed desiderata for improved metrics to enhance interpretability and comparability of existing metrics across datasets of different sizes and properties. Similarly, Hubert et al.~\cite{hubert2023sem} proposed \(Sem@K\), a semantic-oriented metric, to evaluate the semantic-awareness of KGE models. Other studies explored different evaluation protocols~\cite{wang-etal-2019-evaluating,ruffinelli2024beyond} and assessed KGE models from an information retrieval perspective~\cite{zhou2022re}. Additionally, KGE models have been evaluated via real-world applications, such as recommendation systems~\cite{heist2023kgreat}. ReliK~\cite{egger2024relik} introduced a framework to quantify a KGE model's reliability for specific downstream tasks without requiring their direct execution. Furthermore, some studies focused on what the embeddings learn and whether similar entities have similar embeddings~\cite{hubert2024similar}. Others examined biases in KGE benchmarks~\cite{sawischa2023bias,akrami-sigmod}. The most closely related work to ours is~\cite{shirvani2023comprehensive}, which investigates the impact of various data modeling idiosyncrasies, such as representing n-ary relations via mediator nodes, on the performance of KGE models at scale. Additionally, they provide several well-processed datasets that facilitate more realistic assessments of KGE models, two of which are used in our experiments. In this paper, we primarily focus on the limitations of evaluation metrics and protocols, rather than solely on data modeling. Nevertheless, all our experiments place emphasis on large-scale evaluation and systematically examine the effect of different data modeling choices on model performance as well.
 
\vspace{-0.4 cm}
\section{Background}\label{sec:background}
\vspace{-0.2 cm}


\subsection{Common Evaluation Protocols and Evaluation Metrics}\label{sec:eval-back}\vspace{-0.2 cm}
KG embedding-based methods consist of two key components: (1) a score function that evaluates the plausibility of triples \triple{h}{r}{t} by assigning a real-valued score based on relation and entity embeddings, and (2) a learning process to generate the representations (i.e., embeddings) of entities and relations. These models are evaluated on several KG completion tasks such as link prediction, triple classification, relation extraction~\cite{weston,TransR}, and relation prediction~\cite{proje}. However, link prediction remains the most widely used. The goal of link prediction is to infer the missing entity \entity{h} or \entity{t} in a triple \triple{h}{r}{t}. For each test triple \triple{h}{r}{t}, the head entity \entity{h} is replaced with every other entity \(\entity{h'} \in \mathcal{E}\) to form \emph{corrupted} triples, where \(\mathcal{E}\) represents the set of all entities in the dataset. 
The original test triple and its corresponding corrupted triples are ranked by their scores according to the score function and the rank of the original test triple is denoted \emph{rank}$_\entity{h}$.  The same procedure is used to calculate \emph{rank}$_\entity{t}$ for the tail entity \entity{t}.  The ideal score function and embeddings should rank the test triple at top.

The commonly used metrics for measuring the accuracy of embedding models are: (1) \texttt{\small Hits@k$^\uparrow$},\footnote{Throughout the paper, we always place an upward (downward, respectively) arrow beside a measure to indicate that methods with greater (smaller, respectively) values by that measure possess higher accuracy.} the percentage of top $k$ results that are correct; (2) Mean Rank (\mr), the mean of the test triples' ranks, defined as \sloppy $
    \frac{1}{2\ |T|}\sum_{\triple{h}{r}{t} \in T}(\emph{rank}_\entity{h} + \emph{rank}_\entity{t})$ in which \(|T|\) is the size of the test set;  and (3) Mean Reciprocal Rank (\mrr)~\cite{yang2022rethinking}, the average inverse of harmonic mean of the test triples' ranks, defined as $
  \frac{1}{2\ |T|}\sum_{\triple{h}{r}{t} \in T}(\frac{1}{\emph{rank}_\entity{h}} + \frac{1}{\emph{rank}_\entity{t}})   
$. 
In addition to these raw metrics, their corresponding \emph{filtered} versions~\cite{bordes2013translating} are commonly used to exclude the corrupted triples that already exist in the training, validation, or test sets. 
 Throughout the paper, all mentions of the metrics refer to their filtered versions.
 
\vspace{-0.3 cm}
\subsection{Evaluation Datasets and Frameworks} \vspace{-0.2 cm}
The most commonly used datasets for evaluating KGE models are derived from knowledge bases such as Freebase~\cite{freebase}, WordNet~\cite{wordnet}, YAGO~\cite{yago}, Wikidata~\cite{wikidata}, NELL~\cite{NELL}, and DBpedia~\cite{DBpedia}. Among them, FB15k, FB15k-237~\cite{bordes2013translating}, WN18, WN18RR~\cite{ConvE}, NELL-995~\cite{NELL}, and YAGO3-10~\cite{YAGO3} are the most widely used benchmarks. However, these datasets have two key limitations. First, they do not capture n-ary relationships among entities. 
Second, they are substantially smaller than real-world datasets. Large-scale datasets have been introduced to address scalability concerns, yet they also have drawbacks. For instance, Freebase86m~\cite{dgl-ke} includes numerous non-subject matter triples from Freebase’s implementation domains and suffers from high data redundancy due to reverse triples~\cite{shirvani2023comprehensive}. WikiKG90Mv2~\cite{hu2021ogb} and ogbl-wikikg2~\cite{hu2020open} do not include n-ary relationships, i.e., statement nodes in Wikidata. JF17k~\cite{wen2016representation} and WikiPeople~\cite{guan2019link}, albeit small in scale, are also commonly used in the literature. However, these datasets explicitly capture hyper-relational structures~\cite{rosso2020beyond} and are primarily used to evaluate hyper-relational models. 

Several frameworks have been developed to scale up the training and evaluation of KGE models, including DGL-KE~\cite{dgl-ke}, 
PyTorch-BigGraph~\cite{pbg}, and Marius~\cite{mohoney2021marius}. While each framework offers distinct advantages, we selected DGL-KE for our experiments due to its ease of use, efficient negative sampling strategies, and high performance. In this study, we trained and evaluated four well-known KGE models---TransE~\cite{bordes2013translating}, DistMult~\cite{DistMult}, ComplEx~\cite{complex}, and RotatE~\cite{RotatE}---using the settings and hyperparameters recommended in~\cite{dgl-ke}. TransE and RotatE are selected as representatives of translational distance models, while DistMult and ComplEx represent semantic matching models~\cite{dgl-ke}. To overcome the limitations of the datasets mentioned above, we employ the FB-CVT-REV and FB+CVT-REV datasets. As mentioned in Section~\ref{sec:intro}, these datasets are large-scale and do not include the data redundancy that exists in FB15k. FB+CVT-REV explicitly represents n-ary relationships using CVT nodes, whereas FB-CVT-REV converts an $n$-ary relationship centered at a CVT node into $n \choose 2$ binary relationships between every pair of entities by concatenating the edges that connect the entities through the CVT node. 


\vspace{-0.3 cm}
\section{Limitations of Current Evaluation Methodologies}\label{sec:issues}\vspace{-0.2 cm}
\subsection{Shortcomings of Standard Evaluation Metrics}\label{sec:metric-issue}
\subsubsection{Closed-World Assumption}\label{sec:closeissue}
\vspace{-0.1 cm}
The aforementioned performance metrics are designed to evaluate a model's performance based on the closed-world assumption. 
Suppose that the total knowledge of a domain is available. Then it is not necessary to explicitly represent negative facts; they are inferred from the absence of their positive counterparts. 
This concept of closed-world assumption, first introduced by Reiter~\cite{reiter1981closed,reiter1978deductive}, can be stated as ``if a positive fact $P$ is not provable from the knowledge base, assume $\neg P$.'' This assumption is far from realistic due to the incompleteness of real-world knowledge graphs and thus has flaws when a model correctly predicts a triple that does not exist in a dataset. More specifically, if a corrupted triple of a given test triple does not exist in the current dataset, it is considered false. However, the corrupted triple might actually be valid. If a model ranks the correct corrupted triple higher than the test triple itself, its accuracy measures will be penalized, which contradicts the primary goal of link prediction---finding correct triples that do not already exist in the dataset.

\vspace{-0.5 cm}
\subsubsection{Global Metrics}\label{sec:gloabl-metricissue}
Another limitation of the current metrics is that they aggregate the accuracy of all predicted triples into a single value---the micro-average accuracy---which masks the specific strengths and weaknesses of the models. The Freebase dataset exhibits a highly skewed distribution, with a few high-frequency relations and a large number of low-frequency ones~\cite{akrami2021efficacy}. Thus, the micro-average accuracy is predominantly influenced by the most frequent relations. Moreover, the practice of reporting a single, overall accuracy neglects important sources of variability, such as asymmetries between predicting head versus tail entities and performance variation across semantic domains, which can all affect a model’s real-world utility. Furthermore, in the presence of n-ary relationships (i.e., CVT nodes), link prediction becomes more challenging since these nodes are often long-tail entities with limited connectivity~\cite{shirvani2023comprehensive}. This limited connectivity stems from the fact that each CVT node typically represents the specific attributes or relationships of just a single main entity, inherently restricting its connections within the graph. However, their impact on the effectiveness of current link prediction models remains underexplored. 

\vspace{-0.5 cm}
\subsection{Inherent Weaknesses in
Evaluation Protocols}\label{sec:protocol-issue}\vspace{-0.2 cm}
Link prediction is a widely used evaluation protocol for assessing KGE models. However, it has certain limitations. \textit{First}, its evaluation setup can be misleading as we cannot verify if a model ranks all types of false or nonsensical triples lower than correct triples. For example, consider a test triple \triple{James Ivory}{director/film}{A Room With A View}.  Evaluation by link prediction only seeks responses to two sensible questions: ``who directed the film \entity{A Room With A View}?'', i.e., \triple{?}{director/film}{A Room With A View}, and ``which film(s) are directed by \entity{James Ivory}?'', i.e.,  \triple{James Ivory}{director/film}{?}. It does not verify if a model would rank false or nonsensical triples, such as \triple{A Room With A View}{director/film}{James Ivory} lower than the correct one. In this triple, the subject and object entities are swapped, resulting in an incorrect direction for the relation.  Wang et al.~\cite{wang-etal-2019-evaluating} discussed this problem and proposed to use an alternative protocol called entity-pair ranking. \textit{Second}, link prediction is based on the assumption that the presence of a particular property on an entity is already known.  The evaluation focuses on whether a model can derive the correct property values. In reality, though, it remains a challenge to determine whether a property is valid for a given entity in the first place. Therefore, we propose the property prediction task as another evaluation protocol. Both entity-pair ranking and property prediction, along with a third alternative protocol, triple classification, are discussed in Section~\ref{sec:expProtocol}.
\vspace{-0.5 cm}
\section{Experiments}\label{sec:experiments}\vspace{-0.2 cm}
In conducting our experiments, we utilized the provided training, validation, and test sets of the FB-CVT-REV and FB+CVT-REV datasets, with the split ratio of 90/5/5. Our experiments were performed on an NVIDIA H100 80GB GPU. 

\vspace{-0.4 cm}
\subsection{Empirical Analysis and Improvements of Evaluation Metrics} 
\vspace{-0.2 cm}
This section presents a series of experiments demonstrating the limitations that render commonly used evaluation metrics (cf. Section~\ref{sec:eval-back}) inadequate for assessing KGE models. For each limitation, we propose a set of recommendations for improving the evaluation process to obtain more realistic results.

\vspace{-0.3 cm}
\subsubsection{Closed-World Assumption}\label{sec:close}
\vspace{-0.1 cm}
To verify whether the conventional metrics operate under the closed-world assumption and to assess the correctness of the predicted triples absent from the current dataset, we conducted two sets of experiments, one on FB+CVT-REV and the other on FB-CVT-REV. For each set of experiments, we created a subset dataset by randomly selecting 10\% of the full dataset. The models were trained on this subset, and the evaluation was performed twice, using the subset or the full dataset as the ground truth, respectively. A corrupted triple for a test instance in the subset was considered correct if it exists in the ground-truth dataset. We then filtered out these valid corrupted triples from the ranked predictions in order to avoid penalizing the models for correctly predicting them over the actual test triple. The full dataset provides a more comprehensive knowledge base, and thus, using it as the ground truth helps better assess whether a model is incorrectly penalized due to the closed-world assumption. 
While FB+CVT-REV and FB-CVT-REV are still far from complete, this method was employed for a more realistic evaluation of the models' performance and to demonstrate the limitations associated with the closed-world assumption.

The results of this experiment are presented in Table~\ref{table:open-world-cvt}. Using the full datasets to filter out correct predictions from the ranked list led to improved performance. The models' \mrr\ increased by varying degrees, ranging from 3.94\% to 6.98\% across different models and datasets, with the performance improvement being slightly lower on the dataset with n-ary relationships. This increase in accuracy demonstrates that models are penalized for predicting correct triples and that metrics based on the closed-world assumption underestimate models' accuracy. This poses an intrinsic challenge in evaluating KGE models---We would ideally need an absolute ground-truth dataset which is impossible to come by given any large-scale knowledge graph, and such a ground-truth dataset will not require completion anyways. 

\begin{table*}[h]
\caption{Link prediction results (micro-averaged) on FB-CVT-REV and FB+CVT-REV, using their subsets vs. the full datasets as ground-truth to filter out correct predictions from the ranked list to avoid penalizing the models for valid predictions}\label{table:open-world-cvt}
\vspace{-0.2 cm}
\centering
\footnotesize
\resizebox{0.8\textwidth}{!}{\begin{tabular}{|c|c|c|c|c|c|c|c|c|}
\hline
         & \multicolumn{4}{c|}{\scriptsize{\textbf{ground-truth: FB-CVT-REV subset}}} & \multicolumn{4}{c|}{\textbf{\scriptsize{ground-truth: FB-CVT-REV}}} \\ \hline
\textbf{Model}    & \mrr     & \mr  & \hitone       & \hitten     & \mrr     & \mr  & \hitone       & \hitten      \\ \hline
TransE   & 0.73     & 74.33     & 0.67     & 0.83     & 0.77 (6.01\%)     &  73.80    & 0.67     & 0.84    \\ \hline
DistMult & 0.68     & 69.93     & 0.63     & 0.78     & 0.72 (6.98\%)      & 69.03     & 0.64     & 0.80   \\ \hline
ComplEx  & 0.69     & 71.83     & 0.65     & 0.78     & 0.73 (6.90\%)      & 71.10     & 0.65     & 0.81   \\ \hline
RotatE   & 0.69     & 132.13     & 0.65     & 0.75     & 0.73 (6.41\%)     & 131.54     & 0.65     & 0.78   \\ \hline
\end{tabular}}
\resizebox{0.8\textwidth}{!}{\begin{tabular}{|c|c|c|c|c|c|c|c|c|}
\hline
         & \multicolumn{4}{c|}{\textbf{\scriptsize{ground-truth: FB+CVT-REV subset}}} & \multicolumn{4}{c|}{\textbf{\scriptsize{ground-truth: FB+CVT-REV}}} \\ \hline
\textbf{Model}    & \mrr     & \mr  & \hitone       & \hitten     & \mrr     & \mr  & \hitone       & \hitten      \\ \hline
TransE   & 0.50     & 98.39     & 0.44     & 0.62     & 0.51 (3.94\%)     & 98.04     & 0.44     & 0.63     \\ \hline
DistMult & 0.43     & 134.29     & 0.38     & 0.55     & 0.45 (4.76\%)     & 133.55     & 0.39     & 0.57     \\ \hline
ComplEx  & 0.42     & 141.86     & 0.36     & 0.52     & 0.44 (4.82\%)     & 140.00     & 0.37     & 0.54     \\ \hline
RotatE   & 0.40     & 209.93     & 0.35     & 0.47     & 0.41 (4.10\%)     & 208.66     & 0.35     & 0.50     \\ \hline
\end{tabular}}
\vspace{-0.4 cm}
\end{table*}

\vspace{-0.5 cm}
\subsubsection{Macro-Averaging by Relation}  To establish a metric that provides equal weight to all relations in the evaluation, we propose incorporating the macro-average of per-relation accuracy. The macro-averaged \mrr is defined as $
  \frac{1}{|R|} \sum_{r \in R}(\frac{1}{2\ |T_r|}\sum_{\triple{h}{r}{t} \in T_r}(\frac{1}{\emph{rank}_\entity{h}} + \frac{1}{\emph{rank}_\entity{t}}))   
$, where \(|R|\) is the total number of distinct relations and \(|T_r|\) is the size of the test triples in relation $r$. Table~\ref{table:macro-avg-cvt} compares micro- and macro-averaged \mrr. As observed, \mrr\ improves across all models with macro-averaging,  indicating that treating all relations equally results in higher overall performance. This suggests that models perform relatively well even on low-frequency relations. 

\vspace{-0.6 cm}
\subsubsection{Left vs. Right Entity Link Prediction} We propose evaluating left and right entity link prediction---\triple{?}{r}{t} and \triple{h}{r}{?}, respectively---separately for each relation, instead of relying solely on the overall average accuracy of test triples. Table~\ref{table:macro-avg-cvt} shows that models have considerably higher accuracy on right entity link predictions than left entity link predictions. This distinction reveals important asymmetries in model performance; thus, examining predictions in both directions separately provides a more nuanced understanding of model capabilities.

\vspace{-0.6 cm}
\subsubsection{N-ary vs. Binary Relations} To investigate the impact of n-ary relationships on link prediction, we categorized relations into two types---binary relations, which connect two regular entities, and n-ary (or concatenated) relations. In FB+CVT-REV, n-ary relations link regular entities to CVT nodes, while in FB-CVT-REV concatenated relations represent binary conversions of n-ary relations. Table~\ref{table:cvt} presents the results for both relation types, with the ``all'' columns aggregating them. The results indicate that evaluation on concatenated relations yields overestimated accuracy metrics for most models compared to their performance on n-ary ones. This may result from decomposing n-ary relations into several binary relations through Cartesian product-like conversions. The decomposition allows models to more easily learn these simplified patterns when evaluated separately, creating an illusion of higher capability of the models. Moreover, across all models and datasets, the accuracy on concatenated/n-ary relations is significantly higher than binary relations because they exhibit more consistent structural patterns with lower variability, making them easier for models to learn despite their origin in complex relationships.

\begin{table*}[]
\caption{Link prediction accuracy measured by micro-average vs. macro-average on FB-CVT-REV and FB+CVT-REV}\label{table:macro-avg-cvt}
\vspace{-0.2 cm}
\centering
\footnotesize
\resizebox{0.8\textwidth}{!}{\begin{tabular}{|c|c|c|c|c|c|c|}
\hline
         {\scriptsize \textbf{FB-CVT-REV}} & \multicolumn{3}{c|}{\textbf{Micro-average}} & \multicolumn{3}{c|}{\textbf{Macro-average}} \\ \hline
\textbf{Model}    & Left \mrr       & Right \mrr   & Overall \mrr     & Left \mrr        & Right \mrr & Overall \mrr    \\ \hline
TransE  & 0.46           & 0.74            & 0.67           & 0.48  & 0.77           & 0.69                  \\ \hline
DistMult & 0.51           & 0.85            & 0.70           & 0.52   & 0.89       & 0.74          \\ \hline
ComplEx  & 0.56           & 0.87            & 0.71           & 0.59   & 0.90                  & 0.74          \\ \hline
RotatE   & 0.58           & 0.90            & 0.80           & 0.60   & 0.92                 & 0.82          \\ \hline
\end{tabular}}
\resizebox{0.8\textwidth}{!}{\begin{tabular}{|c|c|c|c|c|c|c|}
\hline
        {\scriptsize \textbf{FB+CVT-REV}}  & \multicolumn{3}{c|}{\textbf{Micro-average}} & \multicolumn{3}{c|}{\textbf{Macro-average}} \\ \hline
\textbf{Model}    & Left \mrr       & Right \mrr   & Overall \mrr     & Left \mrr        & Right \mrr & Overall \mrr    \\ \hline
TransE   & 0.35           & 0.65            & 0.57           & 0.38  & 0.68           & 0.60                  \\ \hline
DistMult & 0.41           & 0.73           & 0.61           & 0.43   & 0.75                   & 0.63          \\ \hline
ComplEx  & 0.45           & 0.78            & 0.62           & 0.47   & 0.80                  & 0.65          \\ \hline
RotatE   & 0.49           & 0.83            & 0.73           & 0.50   & 0.85                 & 0.74          \\ \hline
\end{tabular}}
\vspace{-0.4 cm}
\end{table*}

\begin{table}[]
\caption{Micro-averaged \mrr on FB-CVT-REV and FB+CVT-REV, binary vs. n-ary/concatenated relations}\label{table:cvt}
\vspace{-0.2 cm}
\setlength{\tabcolsep}{8pt}
\centering
\footnotesize
\resizebox{0.8\textwidth}{!}{\begin{tabular}{|l|c|c|c|c|c|c|}
\hline

&  \multicolumn{3}{c|}{\textbf{FB-CVT-REV}} & \multicolumn{3}{c|}{\textbf{FB+CVT-REV}} \\ \hline                    
\textbf{Model}    & \textbf{binary} & \textbf{concatenated} & \textbf{all}& \textbf{binary} & \textbf{n-ary} & \textbf{all}  \\ \hline
\textbf{TransE}  & 0.60   & 0.90   & 0.67    & 0.57   & \textbf{0.96}   & 0.57   \\ \hline
\textbf{DistMult} & 0.64   & 0.89   & 0.70    & 0.61   & 0.77   & 0.61    \\ \hline
\textbf{ComplEx}  & 0.66   & 0.90   & 0.71    & 0.62   & 0.80   & 0.62    \\ \hline
\textbf{RotatE}   & \textbf{0.76}   & \textbf{0.92}  & \textbf{0.80}    & \textbf{0.73}   & 0.88   & \textbf{0.73}    \\ \hline
\end{tabular}}
\vspace{-0.5 cm}
\end{table}

\vspace{-0.5 cm}\subsubsection{Macro-Averaging by Domain}Freebase is a cross-domain common fact KG that comprises factual information in a broad range of domains~\cite{shirvani2023comprehensive}, such as music, sports, biology, etc. We analyzed link prediction performance on the 12 most frequent domains in FB-CVT-REV, which together account for nearly 95\% of all subject-matter triples---those representing real-world facts, as opposed to metadata that describes the data itself~\cite{shirvani2023comprehensive}. Table~\ref{table:domain} reports the number of distinct relations, triples, and \mrr\ for each domain. The results show significant variance in models' accuracy across domains, with overall micro-averaged accuracy heavily influenced by dominant domains, particularly \rel{music}, which accounts for over 60\% of the triples. This highlights the importance of incorporating macro-average evaluation alongside micro-average evaluation to give equal weight to each domain, a factor largely overlooked in the existing literature. Here, the macro-average is defined at the domain level rather than the relation level. The domain-level macro-averaged \mrr is $
  \frac{1}{|D|} \sum_{d \in D}(\frac{1}{2\ |T_d|}\sum_{\triple{h}{r}{t} \in T_d}(\frac{1}{\emph{rank}_\entity{h}} + \frac{1}{\emph{rank}_\entity{t}}))   
$, where \(|D|\) is the total number of distinct domains and \(|T_d|\) is the size of the test triples in domain $d$. Given the large variance of performance across different domains, it is also important to understand the models' performance on a domain is influenced by each domain's characteristics. For instance, the models had almost perfect accuracy on domain \rel{measurement\_unit}. 
Most of the frequent relations in this domain are \textit{1-to-n} relations---relations in which a single head entity can be associated with multiple tail entities.
One example is  \rel{/measurement\_unit/dated\_integer/source-/location/hud\_foreclosure\_area/estimated\_number\_foreclosures}, which is a concatenated property. For instance, when the subject is \entity{ United States Department of Housing and Urban Development}, the objects are \entity{Lake Barcroft}, \entity{Van Vleck}, \entity{Fort Ahby}, and so on. Given the similarity between the objects, it could be easier to predict the subject. 
On the other hand, the models had particularly low accuracy on domains such as \rel{book}. One of the most prevalent relations in this domain is \rel{/book/book\_edition/isbn}. An ISBN value mostly appears in one triple only, with highly rare exceptions, and that triple would be an instance of \rel{/book/book\_edition/isbn}. To predict either the book or the ISBN value, without the ISBN value being connected to anything else, the model would be helpless. 
\begin{table*}[]
\vspace{-0.6 cm}
    \caption{Link prediction performance (\mrr) on the most frequent subject-matter domains in FB-CVT-REV}\label{table:domain}
    \centering
    \small
        \resizebox{0.87\textwidth}{!}{\begin{tabular}{|l|c|c|c|c|c|c|} 
            \hline
\textbf{Domain}&\textbf{\#Triples}&\textbf{\#Relations}&\makecell{\textbf{TransE} \\ \textbf{\mrr}}&\makecell{\textbf{DistMult} \\ \textbf{\mrr}}&\makecell{\textbf{ComplEx} \\ \textbf{\mrr}}&\makecell{\textbf{RotatE} \\ \textbf{\mrr}}\\ \hline
            {/music/} & 76,853,315 & 119 & 0.752 & 0.674 & 0.694  & 0.819\\ \hline
            {/film/} &  8,099,424 & 113 & 0.943 & 0.788 & 0.815  & 0.871\\ \hline
            {/people/} & 7,737,375 & 71 & 0.853 & 0.638 & 0.652  & 0.659\\ \hline
            {/tv/} & 5,276,787 & 148 & 0.970 & 0.922 & 0.929  & 0.945\\ \hline
            {/book/} & 5,258,889 & 90 & 0.590 & 0.284 & 0.268  & 0.290\\ \hline
            {/measurement\_unit/} & 4,496,699 & 239 & \textbf{0.999} & \textbf{0.994} & \textbf{0.995} & \textbf{0.990}\\ \hline
            {/location/} & 3,198,389 & 157 & 0.900 & 0.791 & 0.810 & 0.835\\ \hline
            {/award/} &  3,112,556 & 60 & 0.995 & 0.981 & 0.984  & 0.987\\ \hline
            {/biology/} & 1,459,238 & 71  & 0.814 & 0.596 & 0.606 & 0.560\\ \hline
            {/organization/} & 1,223,205 & 55 & 0.883 & 0.710 & 0.734  & 0.766\\ \hline
            {/education/} &  1,083,133 & 82 & 0.866 & 0.740 & 0.658  & 0.696\\ \hline
            {/sports/} & 996,040 & 105 & 0.973 & 0.871 & 0.902  & 0.910\\ 
            \hline
        \end{tabular}}
        \label{table:domains}
        \vspace{-0.6 cm}
        \end{table*}

\vspace{-0.4 cm}
\subsection{Mitigating Weaknesses in Evaluation Protocols}\label{sec:expProtocol}
\vspace{-0.1 cm}
In this section, we reproduce link prediction results and present the results from alternative evaluation protocols, including entity-pair ranking, property prediction, and triple classification. Our results indicate that the models perform poorly on these alternative evaluations, and there is some mismatch between link prediction accuracy and performance on other tasks. For instance, while RotatE achieves the highest link prediction scores, DistMult and ComplEx excel in triple classification. These findings highlight the need for better training and evaluation strategies for KGE models. 

\vspace{-0.2 cm}
\subsubsection{Link Prediction }\label{sec:lp}
\vspace{-0.2 cm}
We conducted link prediction experiments using the four aforementioned KGE models in the DGL-KE framework on datasets FB-CVT-REV, FB+CVT-REV, and FB15k-237. Table~\ref{table:linkprediction} presents the results, revealing a significant performance gap between large- and small-scale datasets. On the FB15k-237 dataset, the performance differences between models are mostly marginal. However, on the larger datasets, their performances diverge more significantly, and the ranking of models differs from that observed on the smaller FB15k-237. For instance, ComplEx performs the worst on small-scale data but surpasses TransE and DistMult on larger datasets. This highlights that results from small-scale data may not provide a realistic evaluation of the models. Although we originally reported these results 
in~\cite{shirvani2023comprehensive}, they are reproduced here to emphasize the importance of experimenting with large-scale datasets rather than small-scale ones. Additionally, these results are included to provide a basis for comparing the performance of KGE models on link prediction with other evaluation protocols. 

\begin{table*}[h]
\vspace{-0.2 cm}
\caption{Link prediction results (micro-averaged) on FB15k-237, FB-CVT-REV, and FB+CVT-REV}\label{table:linkprediction}
\vspace{-0.2 cm}
\centering
\footnotesize
\resizebox{\textwidth}{!}{\begin{tabular}{|c|c|c|c|c|c|c|c|c|c|c|c|c|}
\hline
         & \multicolumn{4}{c|}{\textbf{FB15k-237}} & \multicolumn{4}{c|}{\textbf{FB-CVT-REV}} & \multicolumn{4}{c|}{\textbf{FB+CVT-REV}}\\ \hline
\textbf{Model}    & \mrr   & \mr    & \hitone            & \hitten    & \mrr   & \mr    & \hitone            & \hitten & \mrr   & \mr    & \hitone            & \hitten      \\ \hline
\textbf{TransE}  & \textbf{0.24}        & \textbf{257.75}           & 0.14   & \textbf{0.44}       & 0.67   & \textbf{48.49}   & 0.61  & 0.78     &  0.57  & \textbf{36.12}  & 0.49  & 0.75  \\ \hline
\textbf{DistMult} & \textbf{0.24}        & 385.12           & 0.14    & 0.43   & 0.70   & 70.49   & 0.66    & 0.77   & 0.61  & 81.84  & 0.56  & 0.70    \\ \hline
\textbf{ComplEx}   & 0.23        & 425.38           & 0.14    & 0.42     & 0.71   & 67.74   & 0.68  & 0.78  & 0.62  & 83.20  & 0.57 & 0.70    \\ \hline
\textbf{RotatE}   & \textbf{0.24}        & 288.43  & \textbf{0.16}  & 0.42      & \textbf{0.80}   & 75.72  & \textbf{0.78}  & \textbf{0.84}  & \textbf{0.73} & 68.43  & \textbf{0.69}   & \textbf{0.80}  \\ \hline
\end{tabular}}
\vspace{-0.6 cm}
\end{table*}

\vspace{-0.3 cm}
\subsubsection{Entity-Pair Ranking }\label{sec:entpair-ranking}
\vspace{-0.2 cm}
Wang et al.~\cite{wang-etal-2019-evaluating} discussed the first limitation with link prediction, mentioned in Section~\ref{sec:protocol-issue} and proposed to use an alternative protocol called entity-pair ranking. Following this evaluation protocol, we would create all possible combinations of entities for a relation and then rank them based on their scores. Specifically, Wang et al.~\cite{wang-etal-2019-evaluating} proposed using Weighted Mean Average Precision (\texttt{\small WMAP@K$^\uparrow$}) and Weighted Precision (\texttt{\small WP@K$^\uparrow$}). 
To treat each relation equally, we employ macro-averaging, which corresponds to reporting the unweighted version of these metrics.

Specifically, the unweighted version of $WMAP@K$ is defined as $MAP@K = \sum_{r \in \mathcal{R}} AP_r@K$, where $R$ is the set of all relations in the test set and $AP_r@K$ is defined as $AP_r@K =\frac{1}{NP_r} \sum_{k \in {K}} P_r@k \times relv_r[k]$ in which $NP_r$ is the lesser of $K$ and $|T_r|$ (the number of test triples for relation $r$), $P_r@k$ is the precision among top $k$ predictions for relation $r$, and $relv_r[k]$ is a Boolean indicator of whether the $k$th prediction is correct (1) or not (0). 
Furthermore, we define the unweighted version of $WP@K$ as $P@K = \sum_{r \in \mathcal{R}} P_r@K$ where $R$ is the set of all relations in the test set and $P_r@K$ is the precision among top $K$ predictions for relation $r$. 

Tables~\ref{table:epr-fb} shows the results. The overall observation is that the models have low performance on entity-pair ranking, especially when CVT nodes exist in the data. DistMult is the worst-performing model, while ComplEx has the best \(P@100\) and RotatE has the best \(MAP@100\). \vspace{-0.2 cm}

\begin{table}[]
\centering
\begin{minipage}{0.48\textwidth}
\vspace{-0.2 cm}
    \caption{Entity-pair ranking results}\label{table:epr-fb}
    \vspace{-0.3 cm}
    \setlength{\tabcolsep}{12pt}
    \centering
    \footnotesize
    \resizebox{1.1\textwidth}{!}{%
        \begin{tabular}{|l|l|l|l|l|}
        \hline
        & \multicolumn{2}{l|}{\textbf{FB-CVT-REV}} & \multicolumn{2}{l|}{\textbf{FB+CVT-REV}}\\ \hline
        \textbf{Model}    & \textbf{\map}    & \textbf{\phund}  & \textbf{\map}    & \textbf{\phund}   \\ \hline
        TransE  & 0.12 & 15.6 & 0.07 & 10.4\\ \hline
        DistMult & 0.11  & 7.7 & 0.04 & 1.1\\ \hline
        ComplEx & 0.17 & \textbf{17.0} & 0.11 & \textbf{14.7}\\ \hline
        RotatE   & \textbf{0.20} & 15.4 & \textbf{0.13} & 11.2 \\ \hline
        \end{tabular}
    }
\end{minipage}
\hfill
\begin{minipage}{0.48\textwidth}
\vspace{-0.2 cm}
    \caption{Property prediction results}\label{table:propertyfb}
    \vspace{-0.3 cm}
    \setlength{\tabcolsep}{12pt}
    \centering
    \footnotesize
    \resizebox{.96\textwidth}{!}{%
        \begin{tabular}{|l|l|l|l|l|}
        \hline
        & \multicolumn{2}{l|}{\textbf{FB-CVT-REV}} & \multicolumn{2}{l|}{\textbf{FB+CVT-REV}}\\ \hline
        \textbf{Model}        & \textbf{\mrr}   & \textbf{\mr}   & \textbf{\mrr}   & \textbf{\mr}\\ \hline
        TransE   & 0.10 & 120.1 & 0.07 & 145.1\\ \hline
        DistMult  & 0.27 & 93.1 & 0.26 & 99.8\\ \hline
        ComplEx & 0.33 & 94.5 & 0.31 & 98.8\\ \hline
        RotatE   & \textbf{0.40} & \textbf{80.9} & \textbf{0.37} & \textbf{97.0} \\ \hline
        \end{tabular}
    }
\end{minipage}
\vspace{-0.2 cm}
\end{table}

\vspace{-1.2 cm}\subsubsection{Property Prediction}\label{sec:prop-prediction}
To conduct property prediction, it is imperative to modify the test set. Specifically, we need to restructure the test data to consist only of (\entity{h}, \rel{r}) pairs. A pair (\entity{h}, \rel{r}) is included in the test set only if no triples of the form \triple{h}{r}{t'} exist in the training set. This ensures that the property \rel{r} for entity \entity{h} is not already observed during training, which preserves the integrity of the evaluation. Given a test case (\entity{h}, \rel{r}), a KGE model is applied to calculate the scores of all such triples \triple{h}{r'}{t'} where $\entity{t'} \in \mathcal{E}$, $\rel{r'} \in \mathcal{R}$, $\mathcal{E}$ is the set of all entities, and $\mathcal{R}$ is the set of all relations. 
The scores are used to evaluate whether the correct relation \rel{r} is assigned a high rank among all candidate relations.   Table~\ref{table:propertyfb} shows that the models have deficient performance on this task and struggle to accurately predict the properties of entities. The \mrr\ of the best-performing model, RotatE, is 0.40 on FB-CVT-REV and 0.37 on FB+CVT-REV. However, when evaluating the models using the link prediction protocol, the \mrr\ of the best-performing model, RotatE, is 0.80 on FB-CVT-REV and 0.73 on FB+CVT-REV. This underscores the need for diverse evaluation protocols as assessing models across different tasks provides a more comprehensive understanding of their strengths and limitations. 

\vspace{-0.4 cm}\subsubsection{Triple Classification}\label{sec:classification}
\vspace{-0.2 cm}
Triple classification is the binary classification of triples regarding whether they are true or false facts. 
To conduct this task, we need to create negative samples for the test and validation sets and then learn a classification threshold for each relation on validation data. After learning the thresholds, each test triple with a score higher than the specific threshold for its relation is considered positive and negative otherwise. Some of the earlier KGE models were evaluated using triple classification~\cite{socher2013reasoning,TransH}. However, their negative triples were generated by randomly corrupting head or tail entities of test and validation triples. Randomly generated negative test cases fail to provide sufficient challenges because they are often semantically implausible triples that models can trivially identify as false, resulting in inflated performance measures. Pezeshkpour et al.~\cite{pezeshkpour2020revisiting} and Safavi et al.~\cite{codex} noted this problem and created some hard negative samples for their benchmark datasets. 
Inspired by their work, for both the test and validation sets of each dataset, we generated a suite of four sets of negative samples. Two of these sets adhere to type constraints—one involving tail entity corruption and the other involving head entity corruption—while the remaining two sets violate these constraints. This distinction is necessary because models tend to perform well in recognizing entity types, making \emph{type-inconsistent} negatives easier to classify and thus offering a less challenging evaluation.

\begin{table*}[]
\vspace{-0.6 cm}
\caption{Triple classification results on FB-CVT-REV}\label{table:class-h}
\vspace{-0.3cm}
\centering
\footnotesize
\resizebox{.66\textwidth}{!}{%
\begin{tabular}{|c|c|c|c|c|c|c|c|c|}
\hline
         & \multicolumn{4}{c|}{\textbf{consistent }\entity{h}} & \multicolumn{4}{c|}{\textbf{inconsistent }\entity{h}} \\ \hline
\textbf{Model}    & Precision  & Recall &  Accuracy   &  F1    & Precision  & Recall  & Accuracy   & F1   \\ \hline
\textbf{TransE}   & 0.56       & 0.60   &  0.57  &  0.60  & 0.87       &  0.75   & 0.80  & 0.79 \\ \hline
\textbf{DistMult} & 0.57       & 0.54   &  0.58  &  0.57  & 0.95       &  0.90   & 0.93  & 0.92 \\ \hline
\textbf{ComplEx}  & 0.60       & 0.54   &  0.59  &  0.57  & 0.96       &  0.93   & 0.94  & 0.94 \\ \hline
\textbf{RotatE}   & 0.60       & 0.58   &  0.58  &  0.58  & 0.94       &  0.89   & 0.92  & 0.92 \\ \hline
\end{tabular}}

\resizebox{.66\textwidth}{!}{%
\begin{tabular}{|c|c|c|c|c|c|c|c|c|}
\hline
         & \multicolumn{4}{c|}{\textbf{consistent }\entity{t}} & \multicolumn{4}{c|}{\textbf{inconsistent} \entity{t}} \\ \hline
\textbf{Model}    & Precision  & Recall &  Accuracy  &  F1   & Precision  & Recall  & Accuracy   & F1   \\ \hline
\textbf{TransE}   & 0.63       & 0.56   & 0.60  & 0.60  & 0.95       & 0.86    & 0.91  & 0.90 \\ \hline
\textbf{DistMult} & 0.65       & 0.59   & 0.61  & 0.62  & 0.96       & 0.93    & 0.94  & 0.94 \\ \hline
\textbf{ComplEx}  & 0.65       & 0.61   & 0.62  & 0.64  & 0.96       & 0.94    & 0.95  & 0.95 \\ \hline
\textbf{RotatE}   & 0.68       & 0.55   & 0.62  & 0.59  & 0.93       & 0.84    & 0.90  & 0.88 \\ \hline
\end{tabular}}
\vspace{-0.7 cm}
\end{table*}

Each entity in Freebase has one or more types and these types are denoted by the special relation \rel{object/type}~\cite{shirvani2023comprehensive}. We used the type information and the ranked list of predictions generated by an embedding model for tail or head entity link prediction to create the aforementioned two sets. To generate a type consistent negative triple for a test triple \triple{h}{r}{t}, we scan the ranked list generated for tail entity prediction to identify the first entity \entity{t'} that shares at least one entity type with \entity{t}. We then verify that the corrupted triple \triple{h}{r}{t'} is negative by checking whether it exists anywhere in the whole dataset. If it does not exist in the dataset, triple \triple{h}{r}{t'} is added to the set of type-consistent negative triples for tail entities. We repeat the same procedure and create another set of type-consistent negative triples by corrupting the head entities of test triples. The same procedure is used to create negative samples for validation data. To generate type-violating negative triples, we just make sure that the types of the entity that is used to corrupt a positive triple do not overlap with the original entity's types. 

The results of triple classification on all these new test sets, for both FB-CVT-REV and FB+CVT-REV, are presented in Table~\ref{table:class-h}. It can be observed that the models have low performance on type-consistent negative test samples, while their performance on type-violating ones is satisfactory. The scores of positive and negative triples differ clearly when negative test samples are type-violating. These results suggest that the models primarily learn basic entity type compatibility, rather than capturing deeper semantic distinctions between entities of the same type. 

\begin{table*}[]
\vspace{-0.6 cm}
\caption{Triple classification results on FB+CVT-REV}\label{table:class-h}
\vspace{-0.3cm}
\centering
\footnotesize
\resizebox{.66\textwidth}{!}{%
\begin{tabular}{|c|c|c|c|c|c|c|c|c|}
\hline
         & \multicolumn{4}{c|}{\textbf{consistent }\entity{h}} & \multicolumn{4}{c|}{\textbf{inconsistent }\entity{h}} \\ \hline
\textbf{Model}    & Precision  & Recall &  Accuracy   &  F1    & Precision  & Recall  & Accuracy   & F1   \\ \hline
\textbf{TransE}   & 0.54       & 0.59   &  0.54  &  0.57  & 0.83       &  0.71   & 0.77  & 0.76 \\ \hline
\textbf{DistMult} & 0.56       & 0.54   &  0.54  &  0.54  & 0.95       &  0.88   & 0.92  & 0.91 \\ \hline
\textbf{ComplEx}  & 0.58       & 0.52   &  0.55  &  0.53  & 0.96       &  0.90   & 0.94  & 0.93 \\ \hline
\textbf{RotatE}   & 0.56       & 0.57   &  0.54  &  0.54  & 0.91       &  0.86   & 0.89  & 0.90 \\ \hline
\end{tabular}}

\resizebox{.66\textwidth}{!}{%
\begin{tabular}{|c|c|c|c|c|c|c|c|c|}
\hline
         & \multicolumn{4}{c|}{\textbf{consistent }\entity{t}} & \multicolumn{4}{c|}{\textbf{inconsistent} \entity{t}} \\ \hline
\textbf{Model}    & Precision  & Recall &  Accuracy  &  F1   & Precision  & Recall  & Accuracy   & F1   \\ \hline
\textbf{TransE}   & 0.61       & 0.57   & 0.59  & 0.58  & 0.91       & 0.83    & 0.88  & 0.87 \\ \hline
\textbf{DistMult} & 0.61       & 0.58   & 0.60  & 0.59  & 0.95       & 0.91    & 0.93  & 0.92 \\ \hline
\textbf{ComplEx}  & 0.64       & 0.60   & 0.62  & 0.62  & 0.96       & 0.91    & 0.95  & 0.94 \\ \hline
\textbf{RotatE}   & 0.65       & 0.52   & 0.61  & 0.57  & 0.89       & 0.80    & 0.86  & 0.84 \\ \hline
\end{tabular}}
\vspace{-0.8 cm}
\end{table*}

\vspace{-0.3 cm}
\section{Conclusion}\label{sec:conclusions}
\vspace{-0.2 cm}
This study highlights critical limitations in current evaluation methodologies for KGE models. Our comprehensive large-scale evaluation revealed significant performance variations between small and large datasets, systematic overestimation when n-ary relations are binarized, and flaws in standard metrics. Current metrics suffer from two key problems: they operate under the closed-world assumption, penalizing models for correctly predicting missing triples, and they typically use micro-averaging, which overemphasizes frequent relations while masking performance differences across relations, relation types, and domains. Our evaluation using alternative protocols—entity-pair ranking, property prediction, and triple classification—showed poor performance across all tested models, revealing a disconnect between link prediction scores and broader KG completion capabilities. These findings emphasize the need for more realistic evaluation frameworks and improved KGE methods that better address real-world knowledge graph completion challenges. 


\vspace{-0.4 cm}

\end{document}